\documentclass[runningheads]{llncs}

\usepackage{eccv}

\usepackage{colortbl}
\usepackage{multirow}
\usepackage[accsupp]{axessibility}
\usepackage{algorithm}
\usepackage{algpseudocode} %
\usepackage{setspace}
\usepackage{colortbl}
\usepackage{graphicx}
\usepackage{animate}
\usepackage{caption}
\usepackage{marvosym}

\usepackage{eccvabbrv}

\usepackage{graphicx}
\usepackage{booktabs}

\usepackage[accsupp]{axessibility}  

\usepackage{hyperref}

\usepackage{orcidlink}
\usepackage{xcolor}

\makeatletter
\providecommand{\maketitlesupplementary}{%
  \par\bigskip
  \begin{center}
    {\Large\bfseries\@title\\[0.5em] Supplementary Material}
  \end{center}
  \bigskip
}
\makeatother

\begin{document}
\title{OmniCamera: A Unified Framework for Multi-task Video Generation with Arbitrary Camera Control} 

\titlerunning{Multi-task Video Generation with Arbitrary Camera Control}

\author{Yukun Wang\inst{1,2}\thanks{Work done during an internship at Tencent Hunyuan.} \and
Ruihuang Li\inst{2}\textsuperscript{\Letter} \and
Jiale Tao\inst{2} \and
Shiyuan Yang\inst{2,3} \and
Liyi Chen\inst{2,4} \and
Zhantao Yang\inst{2} \and
Handz\inst{2} \and
Yulan Guo\inst{1}\textsuperscript{\Letter} \and
Shuai Shao\inst{2} \and
Qinglin Lu\inst{2}}

\authorrunning{Y.~Wang et al.}

\institute{\textsuperscript{1}Sun Yat-sen University \quad
\textsuperscript{2}Hunyuan, Tencent \quad
\textsuperscript{3}CityU \quad
\textsuperscript{4}PolyU}

\maketitle

\begin{center}
    \centering
    \includegraphics[width=1\textwidth]{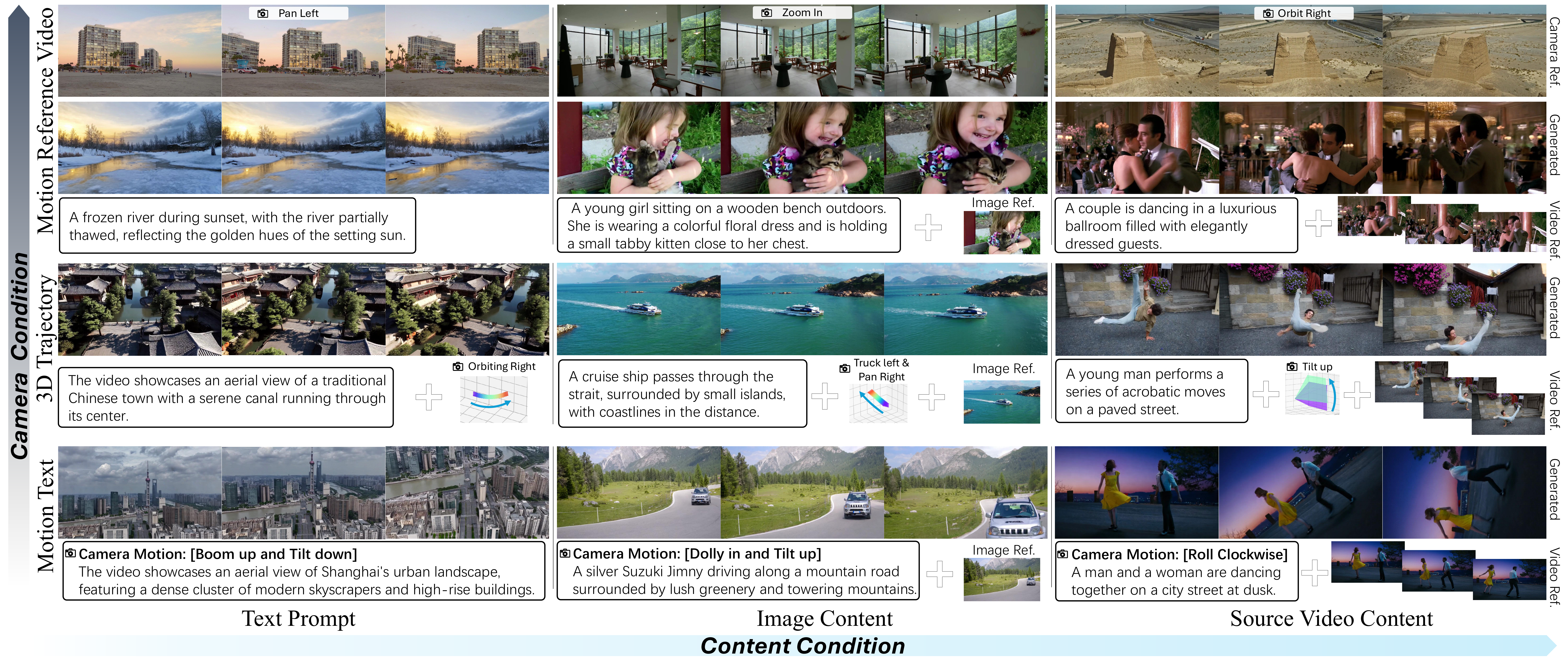}
    \vspace{-1.3em}
    \captionof{figure}{
    We propose \textbf{OmniCamera}, a unified framework that conceptually decouples video generation into two independent control dimensions: camera pose and scene content. It seamlessly integrates three camera conditions (text, 3D trajectory, and motion reference video) with three content conditions (text prompt, image, and source video).}
    \label{fig: teaser}
\end{center}




\begin{abstract}

{
Video fundamentally intertwines two crucial axes: the dynamic content of a scene and the camera motion through which it is observed. However, existing generation models often entangle these factors, limiting independent control. In this work, we introduce OmniCamera, a unified framework designed to explicitly disentangle and command these two dimensions. This compositional approach enables flexible video generation by allowing arbitrary pairings of camera and content conditions, unlocking unprecedented creative control.
To overcome the fundamental challenges of modality conflict and data scarcity inherent in such a system, we present two key innovations. First, we construct OmniCAM, a novel hybrid dataset combining curated real-world videos with synthetic data that provides diverse paired examples for robust multi-task learning. Second, we propose a Dual-level Curriculum Co-Training strategy that mitigates modality interference and synergistically learns from diverse data sources. 
This strategy operates on two levels: first, it progressively introduces control modalities by difficulties (condition-level), and second, trains for precise control on synthetic data before adapting to real data for photorealism (data-level). As a result, OmniCamera achieves state-of-the-art performance, enabling flexible control for complex camera movements while maintaining superior visual quality.
}

\end{abstract}

\vspace{-1em}
\section{Introduction}


Fundamentally, every video is the physical projection of real-world scene content observed through continuous camera poses in 3D space. Driven by this universal mechanism, we propose to conceptually decouple video generation into two independent control dimensions: scene content and camera pose. While simulating realistic cinematographic operations is crucial for professional applications, current approaches typically restrict control to a single modality or focus on narrow tasks as shown in Table.~\ref{Comparison_table}. For instance, existing methods separately utilize textual descriptions \cite{wan2025wan,guo2023animatediff,gao2025seedance}, 3D trajectories \cite{he2024cameractrl,bahmani2025ac3d,li2025realcam,bai2025recammaster}, or reference videos \cite{luo2025camclonemaster} for camera motion. They often struggle with the inherent limitations of each modality (e.g., text is too coarse, trajectories are hard to acquire) and fail to support the free combination of diverse content sources and camera conditions.

\begin{figure}[t] 
  \centering
  \includegraphics[width=1\linewidth]{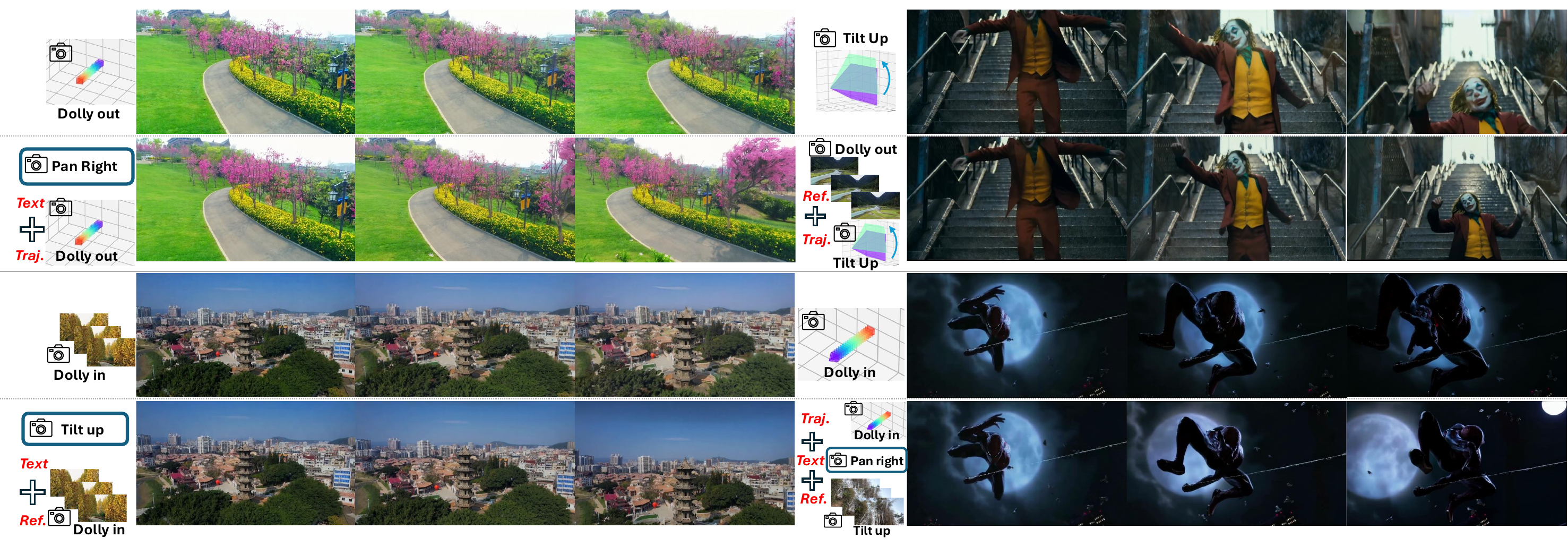} 
  \caption{\textbf{Compositional camera control with multi-modal conditions.} OmniCamera enables flexible combinations of multiple conditioning inputs, such as overlaying text-guided camera motion on top of trajectory or reference-video control, efficiently extending and diversifying camera motion effects.}
  \label{fig:compose}
  \vspace{-1.5em}
\end{figure}

To address these limitations and theoretically encompass all conceivable forms of video generation, we introduce \emph{OmniCamera}, a unified framework that seamlessly integrates these independent conditions. Specifically, as shown in Fig.~\ref{fig: teaser}, the \textbf{camera conditions} involve textual descriptions, explicit trajectory matrices, or reference motion videos, and \textbf{content conditions} can be provided by text prompts, images, or source videos. This framework offers three significant advantages.
Firstly, our synergistic co-training strategy leverages the complementary strengths of diverse data sources. It learns precise camera control from the accurate geometry of synthetic data (e.g., UE5), while simultaneously learning photorealism from real-world videos.
Secondly, as illustrated in Fig.~\ref{fig:compose}, by flexibly combining multiple camera conditions, users can synthesize arbitrarily complex and nuanced camera movements.
Thirdly, our model achieves remarkable parameter efficiency by employing a single set of weights to handle all nine distinct combinations, eliminating the need for specialized models.

Creating a system capable of unified multi-modal camera control is fraught with challenges, foremost among them being the inherent conflict between different control modalities. A naive joint training approach, which simply mixes data from various conditions (e.g., text, reference videos, trajectories), often leads to mutual interference. Besides, there is a profound scarcity of real-world data that simultaneously offers high visual quality and precise camera annotations, making it difficult to train a model that excels in both aspects.

To address these challenges, we present OmniCAM, the first hybrid dataset for camera control, comprising both real-world and synthetic videos and providing diverse types of paired data to facilitate multitask learning. 
As shown in Tab.~\ref{tab:dataset_compare}, OmniCAM is the largest among publicly available datasets and uniquely supports multi-condition training across camera and content dimensions.
The real-world portion of the dataset is meticulously curated through a comprehensive pipeline—including trajectory estimation, filtering, classification, and matching—to select high-quality video pairs with reliable camera trajectories.  

To leverage the unique structure of this dataset, we further propose a dual-level curriculum co-training strategy.
\emph{Condition-level curriculum}. We introduce conditioning modalities in three stages of increasing difficulty. We begin with text-conditioning, the least difficult task, as it closely aligns with the base model's generation ability. Subsequently, we progressively incorporate the more challenging modalities: first, reference-video conditioning, and finally, the most demanding task, trajectory conditioning. This staged approach allows the model to gradually adapt from familiar semantic guidance to complex geometric control.
\emph{Data-level curriculum}. We devise a two-substage training process to combine the strengths of diverse data sources. First, extensive training on large-scale synthetic data (UE5) establishes precise camera control. This is followed by a brief fine-tuning on a small set of real videos to restore photorealism. This short adaptation rapidly leverages the model's inherent visual priors, correcting the realism gap from synthetic data without causing catastrophic forgetting of the learned motion control.
Furthermore, we alleviate modality conflicts using Condition RoPE, which explicitly encodes and separates conditioning inputs.

In summary, our contributions are:
\begin{itemize}
\item We propose \textbf{OmniCamera}, the first unified video generation model that seamlessly integrates three camera conditions (text, trajectory, reference video) with three content conditions (text, image, video). This framework enables arbitrary condition combinations with high parameter efficiency.
\item We curate the \textbf{OmniCAM} dataset, a hybrid dataset combining high-precision synthetic camera trajectories with diverse real-world videos, providing robust multi-level supervision for generalized camera control.
\item We introduce a \textbf{Dual-level Curriculum Co-Training} strategy, comprising a Condition-level Curriculum to avoid modality conflicts and a Data-level Curriculum to bridge the domain gap between synthetic and real data, ensuring stable and effective multimodal learning.
\end{itemize}


Comprehensive quantitative and qualitative experiments demonstrate that our approach surpasses models trained under single-task or single-condition settings, highlighting the effectiveness of the proposed unified framework for camera-controlled video generation.

\begin{table}[t]
\begin{minipage}[b]{0.40\textwidth}
\centering
\caption{Method comparison. OmniCamera decouples video generation into independent content conditions and camera conditions, enabling arbitrary combinations.}
\vspace{-0.5em}
\label{Comparison_table}
\resizebox{\columnwidth}{!}{
\begin{tabular}{l|ccc|ccc}
\hline
\multicolumn{1}{l|}{\multirow{2}{*}{Method}} & \multicolumn{3}{c|}{Content} & \multicolumn{3}{c}{Camera} \\ \cline{2-7} 
\multicolumn{1}{l|}{}                        & Text       & Image      & Video      & Text    & Traj.    & Video    \\ \hline
CameraCtrl~\cite{he2025cameractrl}                                   & \checkmark    &   \checkmark  &  $\times$        &  $\times$       &   \checkmark    &  $\times$            \\
AC3D~\cite{bahmani2025ac3d}    &  \checkmark                        &  $\times$   &  $\times$        &   $\times$      &\checkmark      &  $\times$            \\
RealCam~\cite{li2025realcam}  &   $\times$ & \checkmark   &   $\times$       &   $\times$      & \checkmark   &   $\times$           \\
FloVD~\cite{jin2025flovd}  &    $\times$ & \checkmark   &    $\times$      &    $\times$     & \checkmark   &   $\times$           \\
CamCloneMaster~\cite{luo2025camclonemaster}  & $\times$ & \checkmark   &\checkmark   & $\times$ & $\times$ &  \checkmark    \\
ReCamMaster~\cite{bai2025recammaster}  & $\times$ & $\times$ &   \checkmark & $\times$   & \checkmark   & $\times$ \\
UNIC~\cite{ye2025unic}   & $\times$& $\times$&\checkmark   & $\times$ &\checkmark  & $\times$ \\
OmniVCus~\cite{cai2025omnivcus}  & $\times$& $\times$&\checkmark   & $\times$ &\checkmark  & $\times$\\
Wan-Fun-Camera~\cite{wan2025wan}  & $\times$& \checkmark   & $\times$  & \checkmark & $\times$ & $\times$\\
\hline
\textbf{Ours}                                         & \checkmark  &  \checkmark  & \checkmark  & \checkmark  &  \checkmark & \checkmark   \\ \hline
\end{tabular}}
\end{minipage}
\hfill
\begin{minipage}[b]{0.58\textwidth}
\centering
\caption{Comparison of camera-controllable datasets. OmniCAM is the only dataset providing comprehensive data types for both synthetic and real-world scenarios. \textbf{SS}: Same-Scene, Diverse-Trajectory; \textbf{ST}: Same-Trajectory, Diverse-Scene.}
\vspace{-0.5em}
\label{tab:dataset_compare}
\resizebox{\columnwidth}{!}{
\begin{tabular}{c|l|ccccc|cc}
\hline
\multicolumn{1}{c|}{\multirow{2}{*}{}} & \multicolumn{1}{l|}{\multirow{2}{*}{Dataset}} & \multicolumn{5}{c|}{Data Types} & \multicolumn{2}{c}{Preference} \\ \cline{3-9}
& & Text &Traj.  & SS-Data & ST-Data & Triplet & \#Videos & Resolution \\ \hline
\multirow{4}{*}{\rotatebox{90}{\textit{Syn.}}}
& SynCamMaster~\cite{bai2024syncammaster} & $\times$ &\checkmark &\checkmark & $\times$ & $\times$ & 34K & $1280\!\times\!1280$ \\
& ReCamMaster~\cite{bai2025recammaster} & $\times$ &\checkmark &\checkmark & $\times$ & $\times$ & 136K & $1280\!\times\!1280$ \\
& CamCloneMaster~\cite{luo2025camclonemaster} & $\times$ & $\times$ & $\times$ & \checkmark &\checkmark & 391K & $1008\!\times\!576$ \\
\cline{2-9}
& \textbf{Ours (UE5)} & \checkmark & \checkmark & \checkmark & \checkmark & \checkmark & \textbf{500K} & $\mathbf{1280\!\times\!1280}$ \\ \hline
\multirow{6}{*}{\rotatebox{90}{\textit{Real}}}
& RealEstate-10K~\cite{zhou2018stereo} & $\times$ & \checkmark & $\times$ & $\times$ & $\times$ & 10K & $1280\!\times\!720$ \\
& DL3DV-10K~\cite{ling2024dl3dv} & $\times$ & \checkmark & $\times$ & $\times$ & $\times$ & 10.5K & $3840\!\times\!2160$ \\
& ScanNet~\cite{dai2017scannet} & $\times$ & \checkmark & $\times$ & $\times$ & $\times$ & 1513 & $1296\!\times\!968$ \\
& Matterport3D~\cite{Matterport3D} & $\times$ & \checkmark & $\times$ & $\times$ & $\times$ & 90 & $1280\!\times\!1024$ \\
& ARKitScenes~\cite{dehghan2021arkitscenes} & $\times$ & \checkmark & $\times$ & $\times$ & $\times$ & 1661 & $1920\!\times\!1440$ \\
\cline{2-9}
& \textbf{Ours (Real)} & \checkmark & \checkmark &\checkmark & \checkmark &\checkmark & \textbf{330K} & $\mathbf{3840\!\times\!2160}$ \\ \hline
\end{tabular}}
\end{minipage}
\end{table}

\vspace{-1em}
\section{Related Works}
\vspace{-1em}
\textbf{Video Generation.}
Recent advances in video generation have led to rapid progress, with numerous studies focusing on text-to-video (T2V)~\cite{brooks2024video, hong2022cogvideo, gao2025seedance, wan2025wan, ma2025latte, blattmann2023stable} and image-to-video (I2V)~\cite{xing2024dynamicrafter, yang2024cogvideox, wan2025wan, gao2025seedance} generation. 
Inspired by the success of powerful image generation models such as Stable Diffusion~\cite{esser2024scaling} and Flux~\cite{flux2024}, recent methods have widely adopted Diffusion Transformers (DiT)~\cite{peebles2023scalable} and flow matching~\cite{lipman2022flow} architectures for video generation. 
Modern approaches increasingly aim to unify text and image conditioning within a single framework. 
For example, Wan 2.2~\cite{wan2025wan} introduces a TI2V model that replaces the first-frame latent with image embeddings for I2V tasks, while Seedance~\cite{gao2025seedance} jointly trains T2V and I2V models and integrates video-specific RLHF with multi-dimensional reward mechanisms to enhance quality.

\noindent \textbf{Camera-controlled Video Generation.}
Existing methods incorporate various camera motion control conditions into T2V and I2V models, including textual descriptions~\cite{wan2025wan,guo2023animatediff}, 3D trajectories~\cite{he2024cameractrl,bahmani2025ac3d,li2025realcam,jin2025flovd,he2025cameractrl,bahmani2025ac3d}, and reference videos~\cite{luo2025camclonemaster}.
CameraCtrl~\cite{he2024cameractrl} trains a camera adaptor integrated with T2V and I2V models to inject trajectory-based camera motion, while AC3D~\cite{bahmani2025ac3d} explores camera motion representations within diffusion transformers to achieve controllable generation.
Several studies~\cite{feng2024i2vcontrol,jin2025flovd,li2025realcam,ren2025gen3c,yu2024viewcrafter,xu2024camco,hou2024training} further leverage expert models such as depth~\cite{yang2024depth,ren2025gen3c} and optical flow~\cite{jin2025flovd} networks to provide geometric priors via point clouds or motion fields.
However, the effectiveness of motion control remains bounded by the precision of expert models and the overhead of computing explicit geometric information.

\noindent \textbf{Camera-controlled Video-to-Video Generation.}
Camera-controllable V2V generation aims to synthesize a new video by combining the visual content of a given video with a novel camera motion hint~\cite{bian2025gs,van2024generative,gu2025diffusion,yu2025trajectorycrafter,zhang2025recapture,bai2025recammaster,luo2025camclonemaster}.
Early studies explored this task using Kubric-simulated data~\cite{van2024generative} or by leveraging additional geometric cues such as 3D point tracking~\cite{bian2025gs,yu2025trajectorycrafter}.
Recent works, such as ReCamMaster~\cite{bai2025recammaster}, directly take a content video and a new trajectory as inputs to re-generate videos by training T2V models on carefully curated datasets.
CamCloneMaster~\cite{luo2025camclonemaster} further extends this idea by introducing a video re-shot model capable of guiding both I2V and V2V generation with video hints.


\begin{figure}[t!] 
  \centering
  \includegraphics[width=1\linewidth]{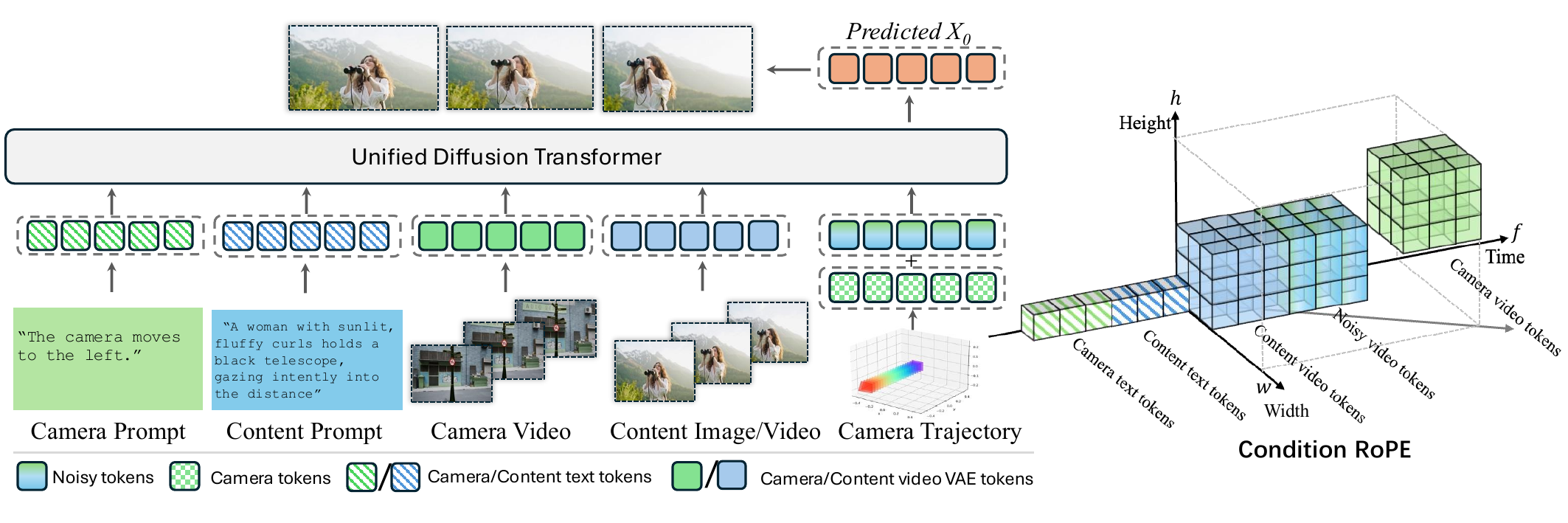} 
  \caption{\textbf{Pipeline of OmniCamera.} Left: Diffusion Transformer with \emph{decoupled} condition injection: text and reference video tokens are concatenated for self-attention; trajectory features are added to the noise latent. Right: Condition RoPE explicitly disentangles modalities using unique positional encodings.}
  \label{fig: pipeline}
  \vspace{-1.5em}
\end{figure}



\section{Method}



Fig.~\ref{fig: pipeline} provides an overview of our OmniCamera framework. We begin by outlining the preliminaries in Sec.~\ref{sec: Preliminary}. Our method is then presented through four key components: \textit{Dataset Design} (Sec.~\ref{sec: Datasetdesign}), \textit{Model Architecture} (Sec.~\ref{sec: ModelDesign}), \textit{Multi-condition Co-Training} (Sec.~\ref{sec: Multi-condition Co-Training}), and \textit{Multi-task Inference} (Sec.~\ref{sec: Multi-task Inference}). 


\vspace{-0.5em}
\subsection{Preliminary}
\label{sec: Preliminary} 
\vspace{-0.5em}
We first introduce the fundamental concept of Flow Matching~\cite{lipman2022flow}, which has been widely adopted in recent video generation models~\cite{wan2025wan}.
Flow Matching formulates generative modeling as learning a continuous velocity field that transports a simple prior distribution (e.g., Gaussian) to the target data distribution.
Given a data sample $x_0 \sim p_{\text{data}}$ and a noise sample $x_1 \sim p_{\text{prior}}$, Flow Matching constructs a linear interpolant:
\begin{equation}
x_t = (1 - t)x_0 + t x_1, \quad t \in [0,1].
\end{equation}
The corresponding target velocity field is defined as:
\begin{equation}
u_t(x_t) = x_1 - x_0,
\end{equation}
which is independent of timestep $t$ but conditioned on the pair $(x_0, x_1)$.
The training objective is to optimize a neural network $v_\theta(x_t, t)$ to approximate this conditional velocity field, thereby recovering the probability flow between the prior and data distributions.

\vspace{-0.5em}
\subsection{Dataset Design}
\label{sec: Datasetdesign} 
\begin{figure}[t]
  \centering
  \includegraphics[width=1\linewidth]{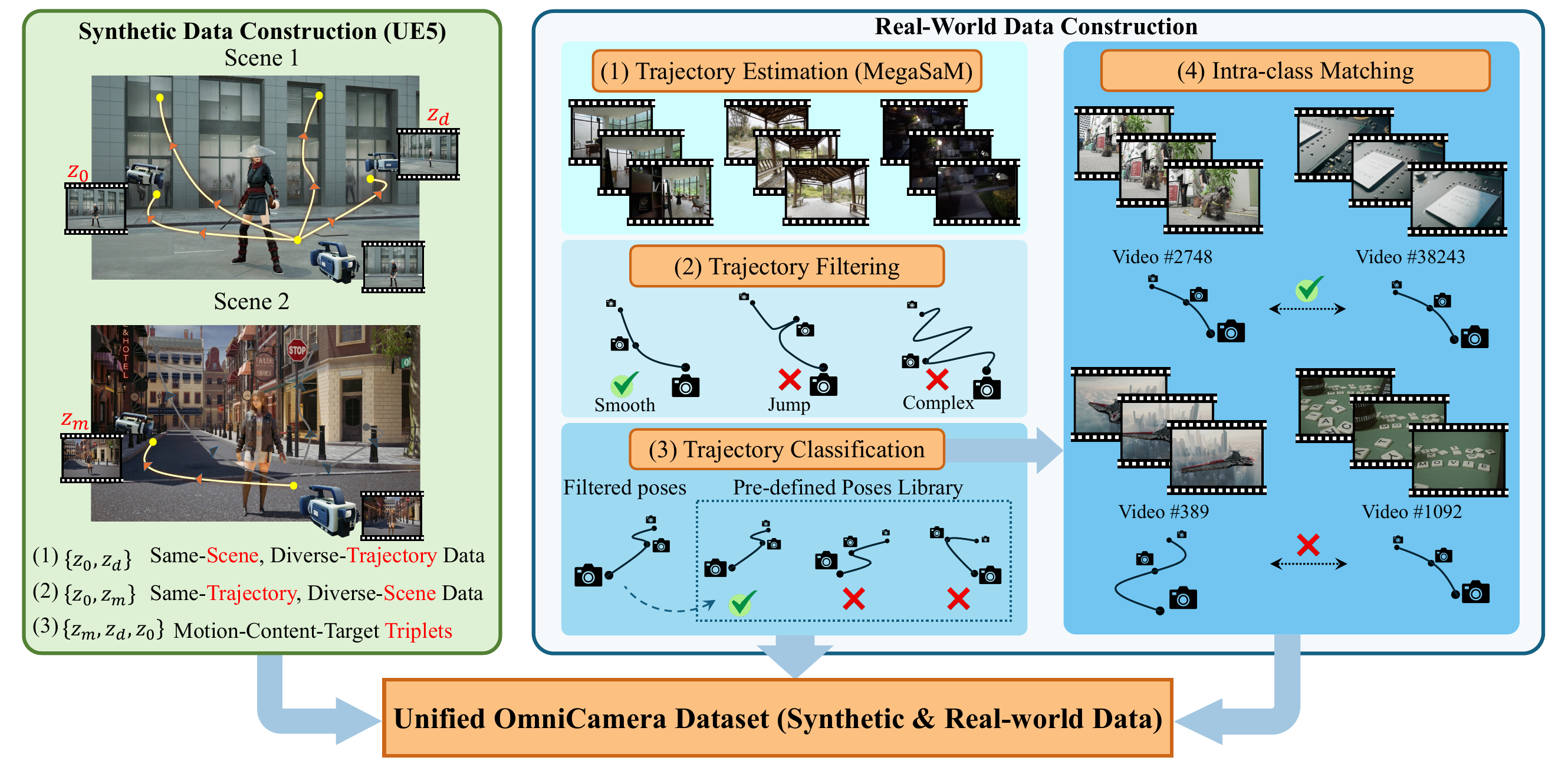}
  \caption{\textbf{OmniCAM dataset construction.} \textbf{Left:} UE5 synthetic videos provide accurate camera poses to build paired and triplet supervision. \textbf{Right:} real videos are processed via trajectory estimation, trajectory filtering, trajectory classification, and intra-class matching to obtain reliable trajectories and cross-scene motion pairs.}
  \label{fig:dataset}
  \vspace{-1.5em}
\end{figure}

\vspace{-0.5em}
Training our unified model requires diverse supervision signals: single videos with trajectories or textual motion annotations, same-scene video pairs with different camera motions, cross-scene pairs with matched motions, and video triplets. We construct the \textbf{OmniCAM} dataset, a hybrid dataset from synthetic and real-world sources, as illustrated in Fig.~\ref{fig:dataset}.

\noindent \textbf{Synthetic Data.}
We define a library of approximately 50 camera-motion types (20 basic and 30 complex), each associated with a detailed textual description, and synthesize corresponding videos in UE5 following~\cite{bai2025recammaster, luo2025camclonemaster}. As shown in the left of Fig.~\ref{fig:dataset}, the synthetic dataset is categorized into three distinct subsets: 

(1) \textbf{Same-Scene, Diverse-Trajectory Data}: Within a fixed \textit{Scene 1}, we first establish a set of camera movement rules to automatically batch-generate diverse camera trajectories. We then simulate a simultaneous shooting process by positioning multiple cameras that face the subjects and move along these predefined trajectories. This allows us to render datasets with synchronized cameras capturing objects from varying perspectives.

(2) \textbf{Same-Trajectory, Diverse-Scene Data}: We place random subjects performing arbitrary actions in \textit{Scene 1} and \textit{Scene 2}. We then render these scenarios by applying identical camera trajectories across these scenes, ensuring consistent camera motion patterns.

(3) \textbf{Motion-Content-Target Triplets}: We construct training triplets denoted by $\{z_m,z_d,z_0\}$, where $z_0$ represents the target video, which shares identical camera motion with the motion video $z_m$ (derived from the same-trajectory, diverse-scene subset) and shares identical content with the content video $z_d$ (derived from the same-scene, diverse-trajectory subset).

In total, we sample 250K distinct camera trajectories (5K trajectories for each of the 50 predefined motion categories) to render 500K independent video clips with accurate pose annotations. By combining these clips, we construct 500K same-scene diverse-trajectory pairs, 500K same-trajectory diverse-scene pairs, and 500K motion-content-target triplets, thereby providing massive and diverse paired supervision.

\noindent \textbf{Real-World Data.}
In addition to synthetic data, we curate a real-world dataset through a rigorous pipeline. As illustrated in the right of Fig.~\ref{fig:dataset}, our data processing pipeline comprises four key steps:

(1) Trajectory Estimation: We utilize MegaSaM~\cite{li2025megasam} to extract camera trajectories, followed by camera parameter calibration from CameraCtrl2~\cite{he2025cameractrl}.

(2) Trajectory Filtering: Since raw trajectories extracted by MegaSaM often contain significant noise, we filter them based on trajectory smoothness. Let $\mathbf{c}_i$ denote the camera position at frame $i$ (where $i \in \{1, \dots, N\}$ for a video of $N$ frames) and $d_i=\|\mathbf{c}_{i+1}-\mathbf{c}_i\|_2$ the frame-to-frame displacement. We compute:
\vspace{-0.5em}
\begin{equation}
r_{\text{jump}} = \frac{\max_i d_i}{\bar d}, \quad r_{\text{comp}} = \frac{L}{\|\mathbf{c}_N-\mathbf{c}_1\|_2+\epsilon},
\end{equation}
where $\bar d=\frac{1}{N-1}\sum_{i=1}^{N-1} d_i$ and $L=\sum_{i=1}^{N-1} d_i$.
We apply two criteria: (i) \textit{jump filtering}, discarding trajectories with $r_{\text{jump}}>\tau_{\text{jump}}$; and (ii) \textit{complexity filtering}, discarding trajectories with $r_{\text{comp}}>\tau_{\text{comp}}$, which indicates overly tortuous or jittery motion. Here, $\tau_{\text{jump}}$ and $\tau_{\text{comp}}$ are preset thresholds, and $\epsilon$ is a small constant for numerical stability.

(3) Trajectory Classification: We categorize the filtered trajectories into $50$ predefined camera motion patterns based on similarity. For each video, we compute the trajectory similarity with all 50 predefined templates based on translation error (TransErr) and rotation error (RotErr)~\cite{he2024cameractrl}, and assign it to the class yielding the minimum error.

(4) Intra-class Matching: Within each class, we perform random pairwise matching of trajectories. We compute the TransErr and RotErr between two candidate trajectories and consider them as a valid match only if both errors are below strict predefined thresholds. This process yields real-world data pairs characterized by identical trajectories but diverse scenes.

Through this progressive pipeline, we obtain data at three levels of annotation: 380K videos with reliable trajectory annotations from Step (2), a total of 380K videos augmented with categorical motion descriptions from Step (3), and 300K Same-Trajectory, Diverse-Scene real-video pairs identified from Step (4).


\vspace{-0.5em}
\subsection{Model Design}
\label{sec: ModelDesign} 
\vspace{-0.5em}
To avoid modality conflicts, we adopt a decoupled condition-injection strategy, as shown in the left of Fig.~\ref{fig: pipeline}: textual prompts and visual conditions (content latent $z_d$ and camera-motion latent $z_m$) are unified as sequence-level representations to interact with the noise latent $z_t$ during joint attention; concurrently, trajectory parameters are processed via an MLP before being passed into DiT blocks.

\noindent \textbf{3D Condition RoPE.} As illustrated in the right of Fig.~\ref{fig: pipeline}, to resolve the spatial-temporal ambiguity caused by sequence concatenation, we propose a 3D Condition RoPE. For an original token coordinate $(f, h, w)$ denoting the frame, height, and width index, we assign distinct spatial-temporal base offsets for each modality: $(0,0,0)$ for $z_t$, $(F,0,0)$ for $z_d$, and $(F,H,W)$ for $z_m$, where $F, H, W$ represent the total frames, height, and width dimensions of the noise latent $z_t$. The positional codes are then computed on these shifted coordinates, with frequencies $\text{freq}_i = \theta^{-2i/D}$, where $D$ is the channel dimension of the token embeddings and $\theta$ is the base constant. This explicitly distinguishes identical coordinate structures, seamlessly integrating multiple conditions within our unified framework.


\begin{figure}[t]
  \centering
  \includegraphics[width=1\textwidth]{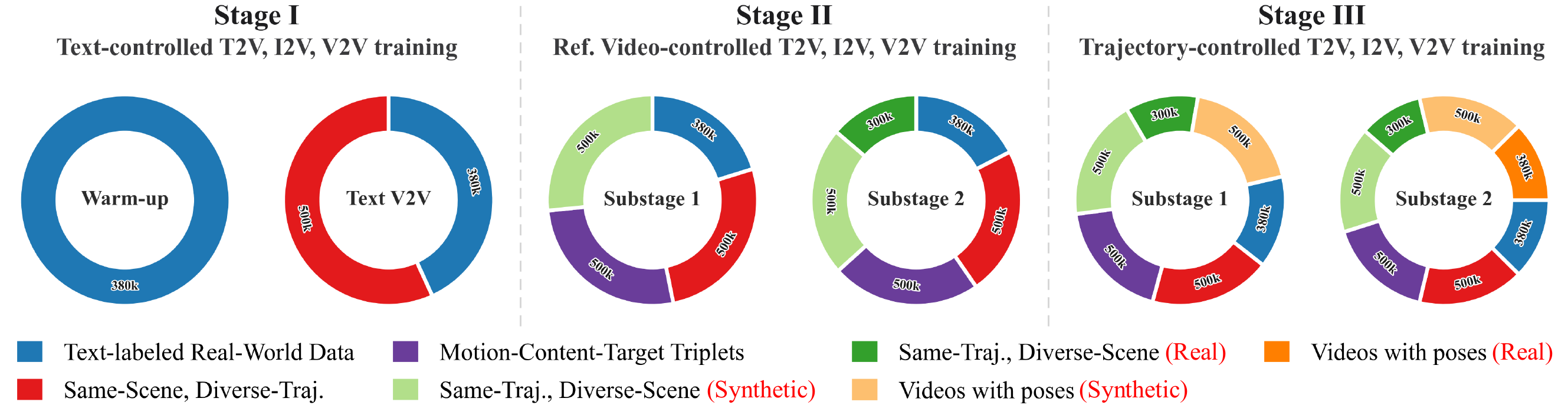}
  \caption{\textbf{Dual-level curriculum data composition.} Condition-level curriculum progresses from text control (Stage~I) to reference-video control (Stage~II) to trajectory control (Stage~III), following a coarse-to-fine and easy-to-hard order. In Stage~II and Stage~III, data-level curriculum first trains on synthetic data with precise trajectories (Substage~1), then fine-tunes on real-world videos to recover photorealism (Substage~2).}
  \label{fig:five_stage_data}
  \vspace{-1.5em}
\end{figure}

\vspace{-0.5em}
\subsection{Dual-level Curriculum Co-Training}
\label{sec: Multi-condition Co-Training} 
\vspace{-0.5em}

Simultaneously training 9 different generation combinations (3 camera-control modalities $\times$ 3 generative tasks) introduces severe representation conflicts and optimization instability. Naively mixing all data and conditions from scratch leads to degraded camera controllability and deteriorated visual quality. To address this, we propose a \textbf{Dual-level Curriculum Co-Training} strategy operating along two complementary axes: a task-level curriculum to align multi-modal conditions progressively, and a data-level curriculum to balance camera controllability and visual realism.

\noindent\textbf{Condition-level Curriculum.}
As shown in Fig.\ref{fig:five_stage_data}, we design a three-stage curriculum that evolves from coarse-to-fine granularity and easy-to-hard difficulty.
In \textbf{Stage I}, we focus on text-based conditioning (across T2V, I2V, and V2V tasks). This acts as semantic guidance with the coarsest control granularity. Since it aligns closely with the base model's inherent capabilities, this stage serves as the foundational and easiest learning step. In \textbf{Stage II}, we incorporate reference-video control. Operating at a relatively coarse granularity, this stage employs in-context learning to transfer coarse global camera motion from a reference video to the target. This introduces a higher level of learning difficulty compared to pure text guidance. In \textbf{Stage III}, we introduce camera trajectory control, which demands the finest granularity and precise geometric control. Consequently, this is the most challenging modality. By adhering to this coarse-to-fine and easy-to-hard progression, we effectively prevent optimization collapse and achieve reliable multi-condition control.

\vspace{0.2em}\noindent\textbf{Data-level Curriculum.}
High-quality multi-condition training faces a severe data dilemma: simulated data (e.g., UE5) provides perfectly accurate trajectories but suffers from a significant domain gap compared to real-world videos; conversely, real-world data possesses high visual realism but yields noisy trajectory estimates via Structure-from-Motion (SfM).
To integrate the advantages of diverse data sources, we develop a two-substage data curriculum applied in the later conditioning stages (Stage~II and Stage~III). In \textbf{Substage 1}, we utilize a large scale of UE5 data paired with precise trajectories to significantly boost camera motion accuracy. In \textbf{Substage 2}, we employ a curated set of high-quality real videos with reliable trajectory estimates to restore the model's capability in generating photorealistic content.

\vspace{-0.5em}
\subsection{Multi-task Inference}
\label{sec: Multi-task Inference} 
\vspace{-0.5em}
We employ a dual-condition classifier-free guidance strategy. The model relies on text $C_T$ for semantic control and multimodal inputs $C_M$ for camera motion. To enable flexible controllability, we randomly drop conditions during training with a probability of 5\% for $C_M$ only, 5\% for $C_T$ only, and 5\% for both simultaneously. At the inference stage, this allows us to utilize separate guidance scales, $w_T$ and $w_M$, to independently adjust the influence of text and motion guidance. The modified score estimate is as follows:
\begin{equation}
    \begin{aligned}
\hat{\epsilon}_\theta\left(z_t, c_{T}, c_{M}\right) & =\epsilon_\theta\left(z_t, \phi, \phi \right) \\
& +w_T\left(\epsilon_\theta\left(z_t, c_T, \phi \right)-\epsilon_\theta\left(z_t, \phi, \phi \right)\right) \\
& +w_M\left(\epsilon_\theta\left(z_t, c_T, c_M\right)-\epsilon_\theta\left(z_t, c_T, \phi \right)\right)
    \end{aligned}
    \label{eq:cfg}
\end{equation}
Here, $c_M \in \{\text{Trajectory}, \text{Ref-Video}\}$ acts as a polymorphic camera condition. 


\vspace{-0.1em}
\section{Experiments}
\vspace{-0.1em}
\subsection{Implementation Details}
\vspace{-0.1em}

We build upon the 5B-parameter Wan2.2-TI2V architecture, utilizing its VAE for $704\times1248$ video generation. During training, we exclusively optimize the self-attention, cross-attention, and camera embedding modules. We train the model on 32 GPUs for 60K steps with a batch size of 32, using the AdamW optimizer (initial learning rate $7\times10^{-5}$ decaying to $5\times10^{-5}$ via a cosine schedule).

\noindent \textbf{Evaluation.} We have constructed a comprehensive evaluation dataset that comprises 36 camera-motion text commands, 36 camera trajectories, and 36 reference videos. For each task, we assess the camera-control performance on 800 generated videos. The evaluation metrics include CLIP-T, CLIP-F, Rotation Error, Translation Error, FVD or FVD-V, as well as motion accuracy. Detailed calculations for these metrics can be found in the Supplementary Material.

\begin{figure}[h] 
  \centering
  \includegraphics[width=1\linewidth]{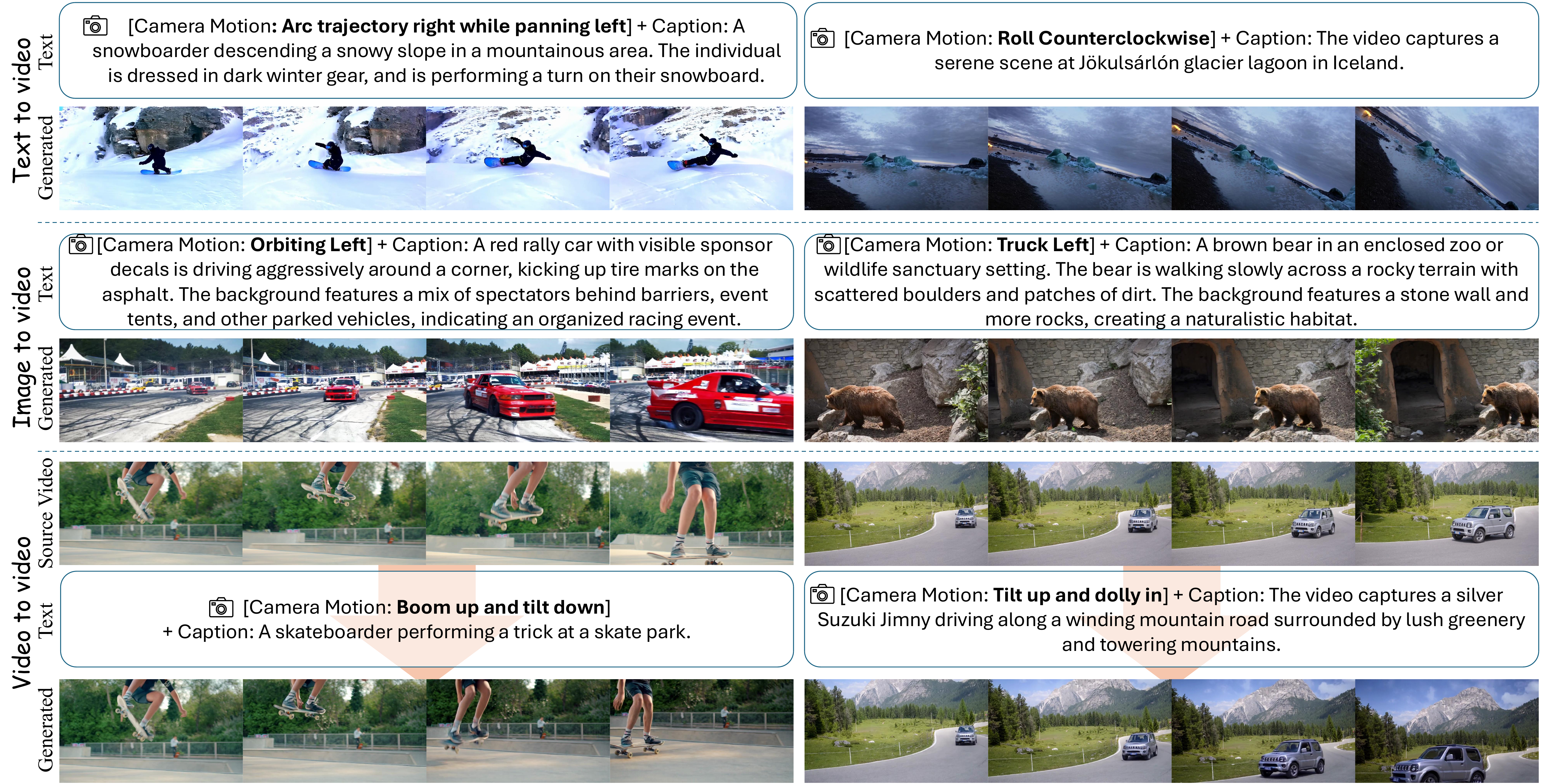} 
  \caption{Qualitative results for \textbf{text-controlled} camera motion. }
  \label{fig:experiment1}
\end{figure}


\vspace{-1em}
\begin{table}[h]
  \centering
  \begin{minipage}[t]{0.49\textwidth}
  \centering
  \caption{Quantitative results of \textbf{Text-controlled} camera motion on T2V, I2V and V2V tasks.}
  \vspace{-0.5em}
  \label{tab:Text-controlled}
  \resizebox{\linewidth}{!}{
  \begin{tabular}{l|l|cccc}
  \toprule
  Task & Method & FVD$\downarrow$ & CLIP-F$\uparrow$ & CLIP-T$\uparrow$ & MotionAcc$\uparrow$ \\
  \midrule
  \multirow{4}{*}{T2V}
   & AnimateDiff~\cite{guo2023animatediff} & 1471.28 & 96.39 & 25.04  & 19.3\% \\
   & Wan2.1-T2V~\cite{wan2025wan} & 899.21 & 98.54 & 27.75  & 28.9\% \\
   & Wan2.2-TI2V~\cite{wan2025wan} & \textbf{867.27} & 98.92 & 28.06  & 31.4\% \\
   & \textbf{Ours} & 884.11 & \textbf{98.94} & \textbf{29.06}  & \textbf{92.4\%} \\
  \midrule
  \multirow{3}{*}{I2V}
   & Wan2.1-I2V~\cite{wan2025wan} & 345.69 & 98.27 & 29.95 & 27.5\% \\
   & Wan2.2-TI2V~\cite{wan2025wan} & \textbf{285.52} & \textbf{99.05} & \textbf{30.30}  & 30.3\% \\
   &Wan2.2-Fun-Camera &314.63 & 99.02 & 30.08 & 68.8\% \\
   & \textbf{Ours} & 308.83 & \textbf{99.05} & 30.29 & \textbf{91.3\%} \\
  \midrule
  \multirow{1}{*}{V2V}
   & \textbf{Ours} & \textbf{430.71} & \textbf{99.02} & \textbf{30.17}  & \textbf{89.7\%} \\
  \bottomrule
  \end{tabular}}
  \end{minipage}\hfill
  \begin{minipage}[t]{0.49\textwidth}
  \centering
  \caption{Quantitative results of \textbf{Trajectory-controlled} camera motion on the tasks (CLIP-T for T2V/I2V, CLIP-V for V2V).}
  \vspace{-0.5em}
  \label{tab:Trajectory-controlled}
  \resizebox{\linewidth}{!}{
  \begin{tabular}{l|l|cccc}
  \toprule
  Task & Method & FVD$\downarrow$ & RotErr$\downarrow$ & TransErr$\downarrow$ & CLIP(T/V)$\uparrow$ \\
  \midrule
  \multirow{3}{*}{T2V}
   & CameraCtrl~\cite{he2024cameractrl} & 1365.90 &0.126 &8.160 &23.43 \\
   & AC3D~\cite{bahmani2025ac3d} & 958.62 &0.056 & 7.542 &\textbf{30.01}  \\
   & \textbf{Ours} & \textbf{893.95} &\textbf{0.034} &\textbf{2.064} &29.64 \\
  \midrule
  \multirow{2}{*}{I2V}
   & CameraCtrl~\cite{he2024cameractrl} & 586.14 &0.048 &3.798 &27.04 \\
   & \textbf{Ours} & \textbf{333.54} &\textbf{0.043} &\textbf{3.355}  &\textbf{30.40} \\
  \midrule
  \multirow{2}{*}{V2V}
   & ReCamMaster~\cite{bai2025recammaster} & 366.82 &0.048 &\textbf{5.320} &94.10 \\
   & \textbf{Ours} & \textbf{331.36} &\textbf{0.045} &5.933 &\textbf{94.97} \\
  \bottomrule
  \end{tabular}}
  \end{minipage}
  \vspace{-0.5em}
  \end{table}
  
  \begin{table}[t]
    \centering
    \caption{Quantitative results of \textbf{Reference-video-controlled} camera motion on T2V, I2V, and V2V tasks. ``-'' means methods fail to accomplish task.}
    \label{tab:Video-controlled}
    \resizebox{1\textwidth}{!}{
    \begin{tabular}{l|cccc|cccc|cccc}
    \toprule
    \multirow{2}{*}{Methods} & \multicolumn{4}{c|}{T2V} & \multicolumn{4}{c|}{I2V} & \multicolumn{4}{c}{V2V} \\
    \cmidrule(lr){2-5} \cmidrule(lr){6-9} \cmidrule(lr){10-13}
     & FVD$\downarrow$ & RotErr$\downarrow$ & TransErr$\downarrow$ &CLIP $\uparrow$  & FVD$\downarrow$ & RotErr$\downarrow$ & TransErr$\downarrow$ &CLIP $\uparrow$ & FVD-V$\downarrow$ & RotErr$\downarrow$ & TransErr$\downarrow$ &CLIP-V $\uparrow$ \\
    \midrule
    CamCloneMaster~\cite{luo2025camclonemaster} &-- &-- &-- &-- &380.73 &\textbf{0.021} &\textbf{3.953} &30.18  &376.33 &\textbf{0.019} &\textbf{4.936} &93.60 \\
    \hline
    Ours    &\textbf{868.63} &\textbf{0.024}  &\textbf{7.109}  &\textbf{29.30} &\textbf{352.76} &0.023 &4.195 &\textbf{30.45} &\textbf{348.51} &\textbf{0.019} &5.392 &\textbf{93.71}  \\
    \bottomrule
    \end{tabular}}
    \vspace{-1em}
    \end{table}
\begin{figure}[t] 
  \centering
  \includegraphics[width=1\linewidth]{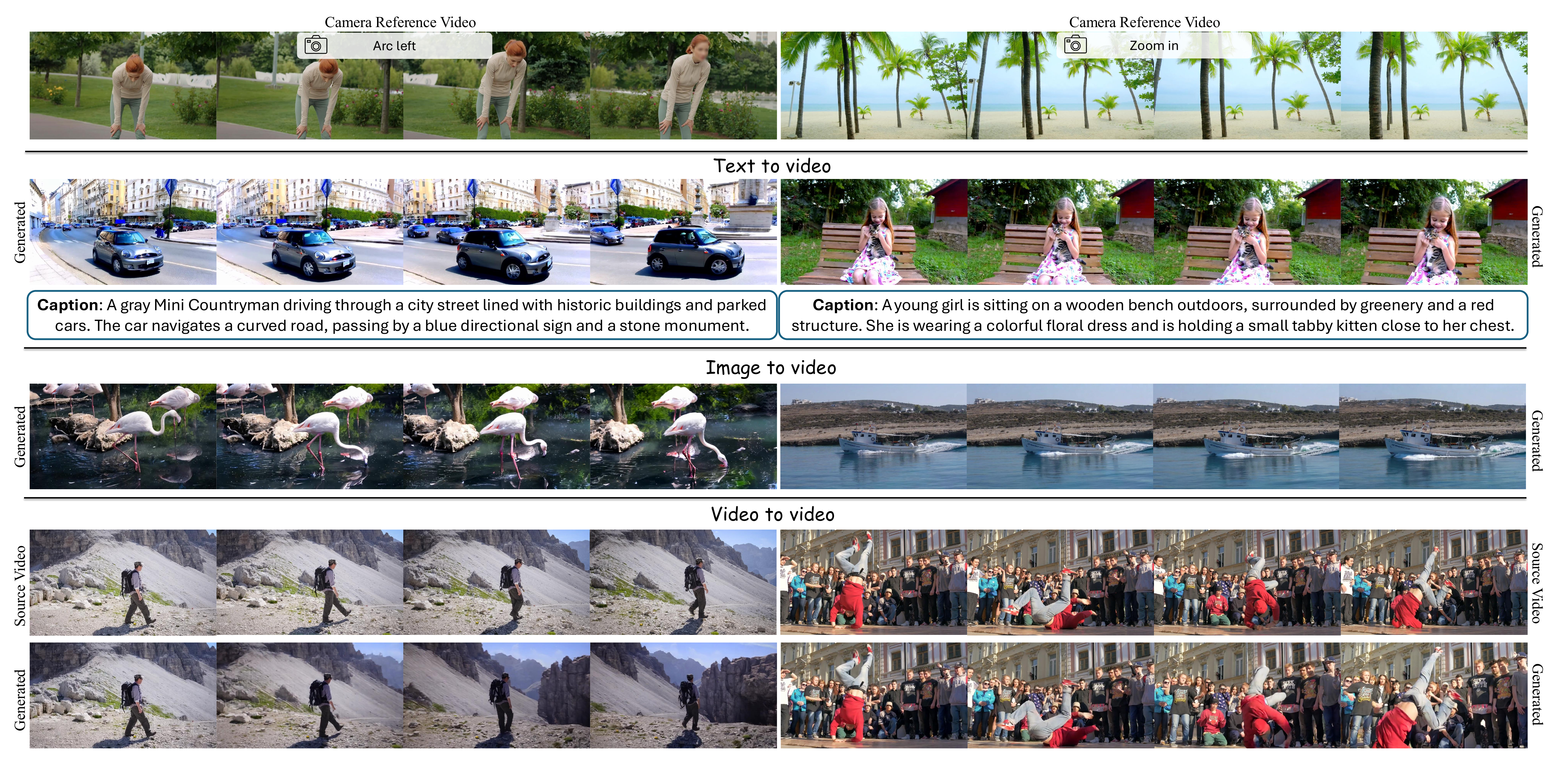} 
  \vspace{-2.0em}
  \caption{Qualitative results for \textbf{reference-video-controlled} camera motion.}
  \label{fig:experiment3}
  \vspace{-1.0em}
\end{figure}


\begin{figure}[t] 
  \centering
  \includegraphics[width=1\linewidth]{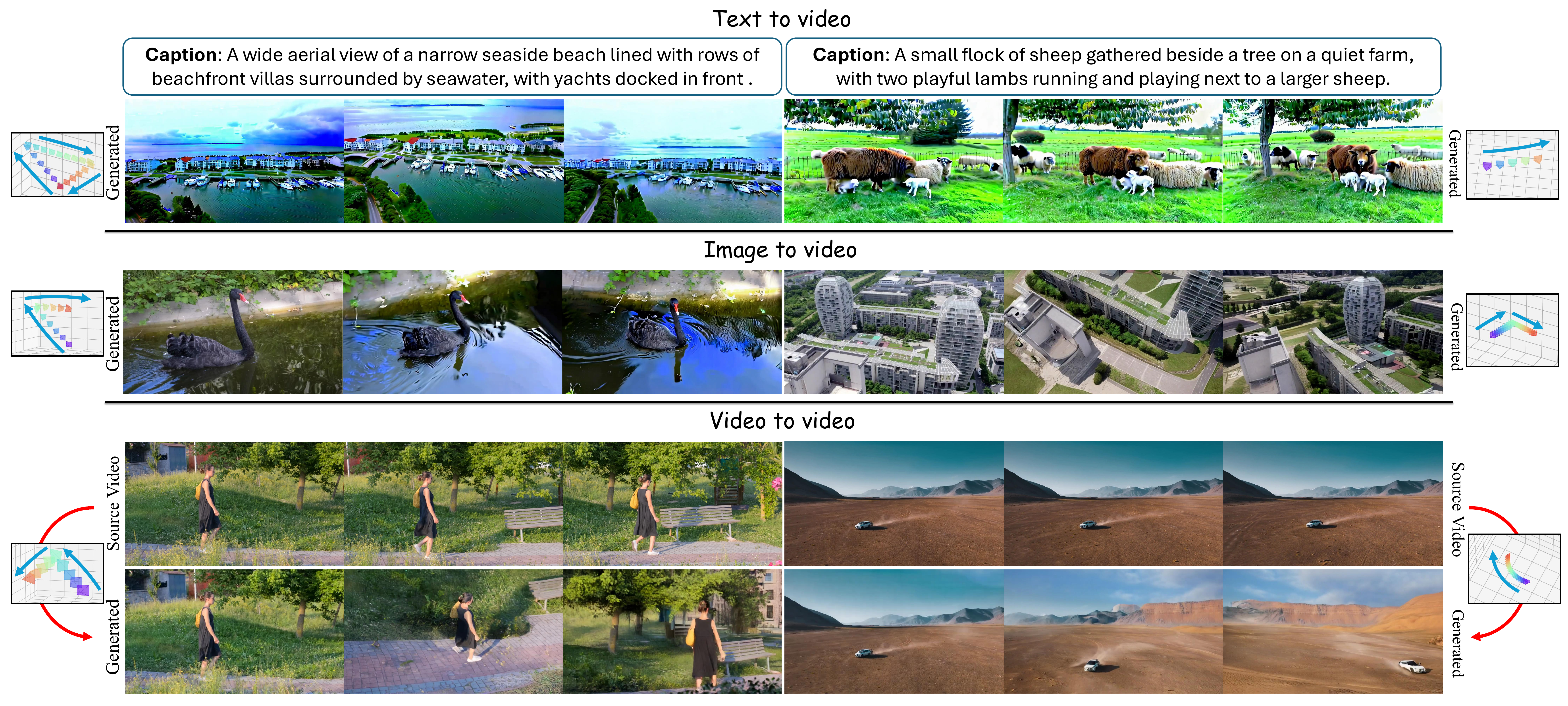} 
  \vspace{-2.0em}
  \caption{Qualitative results for \textbf{trajectory-controlled} camera motion.}
  \label{fig:experiment2}
  \vspace{-1em}
\end{figure}

\begin{figure}[t] 
  \centering
  \includegraphics[width=1\linewidth]{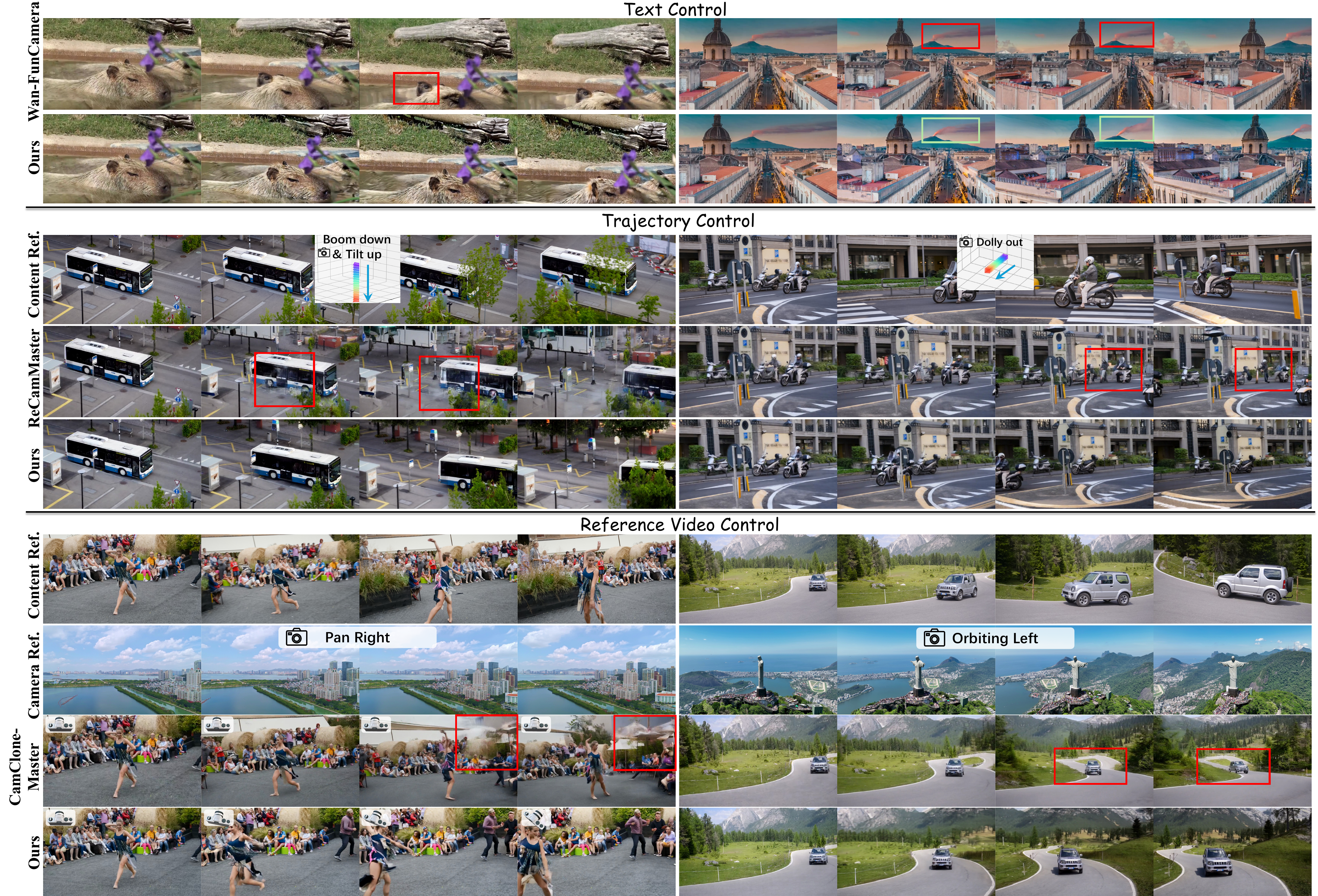} 
  \caption{\textbf{Visual comparison} of OmniCamera with state-of-the-art methods. \textbf{Text Control:} Wan2.2-Fun-Camera produces some artifacts (e.g., the ``ear'' mutating into ``eye''). \textbf{Trajectory Control:} ReCamMaster~\cite{bai2025recammaster} produces severe distortions on the ``bus'' and ``motorcyclist''. \textbf{Reference-Video Control:} CamCloneMaster~\cite{luo2025camclonemaster} executes incorrect camera motion while introducing heavy background and object distortions. }

  \label{fig:experiment4}
  \vspace{-0.6em}
\end{figure}

\begin{figure}[t]
  \centering
  \includegraphics[width=1\linewidth]{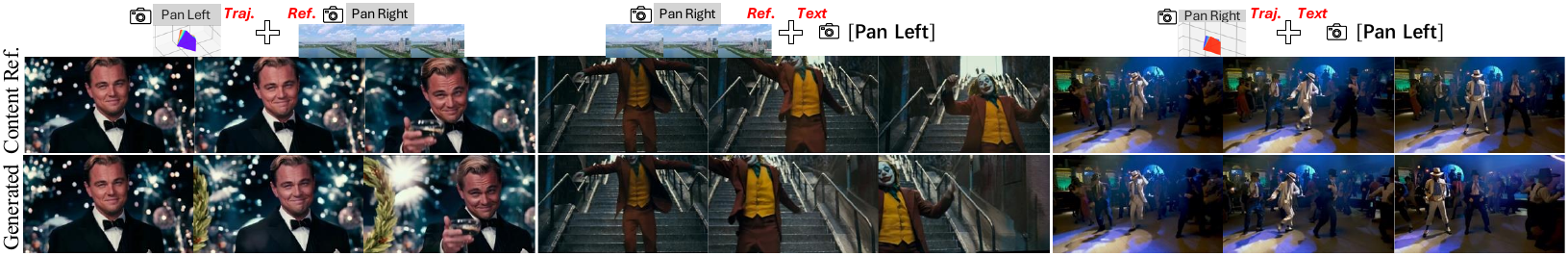}
  \caption{\textbf{Modality conflict analysis.} Pairwise combination of contradictory camera conditions reveals a dominance order: trajectory>reference video>text prompts.}
  \label{fig:mode_conflict}
  \vspace{-0.8em}
\end{figure}

\begin{table}[t]
\centering
\caption{\textbf{Ablation study} of Curriculum Co-Training across different tasks. \textbf{CC}: condition-level curriculum, \textbf{DC}: data-level curriculum.}
\label{tab:ablation_curriculum}
\resizebox{1\textwidth}{!}{
\begin{tabular}{l | c c c | c c c | c c c | c c c | c c c | c c c }
\toprule
\multirow{3}{*}{Method} & \multicolumn{9}{c|}{\textbf{Trajectory-controlled}} & \multicolumn{9}{c}{\textbf{Reference-video-controlled}} \\
\cmidrule(lr){2-10} \cmidrule(lr){11-19}
 & \multicolumn{3}{c|}{T2V} & \multicolumn{3}{c|}{I2V} & \multicolumn{3}{c|}{V2V} & \multicolumn{3}{c|}{T2V} & \multicolumn{3}{c|}{I2V} & \multicolumn{3}{c}{V2V} \\
\cmidrule(lr){2-4} \cmidrule(lr){5-7} \cmidrule(lr){8-10} \cmidrule(lr){11-13} \cmidrule(lr){14-16} \cmidrule(lr){17-19}
 & TransErr$\downarrow$ & RotErr$\downarrow$ & FVD$\downarrow$ & TransErr$\downarrow$ & RotErr$\downarrow$ & FVD$\downarrow$ & TransErr$\downarrow$ & RotErr$\downarrow$ & FVD-V$\downarrow$ & TransErr$\downarrow$ & RotErr$\downarrow$ & FVD$\downarrow$ & TransErr$\downarrow$ & RotErr$\downarrow$ & FVD$\downarrow$ & TransErr$\downarrow$ & RotErr$\downarrow$ & FVD-V$\downarrow$ \\
\midrule
w/o CC & 3.512 & 0.052 & 1024.31 & 4.821 & 0.061 & 412.50 & 7.425 & 0.068 & 405.12 & 9.510 & 0.038 & 982.45 & 5.762 & 0.035 & 410.28 & 7.214 & 0.029 & 415.82 \\
w/o DC & 2.845 & 0.045 & 965.84 & 4.103 & 0.052 & 378.42 & 6.512 & 0.055 & 362.75 & 8.324 & 0.031 & 925.61 & 4.951 & 0.028 & 381.54 & 6.185 & 0.024 & 382.46 \\
\midrule
\textbf{Ours} & \textbf{2.064} & \textbf{0.034} & \textbf{893.95} & \textbf{3.355} & \textbf{0.043} & \textbf{333.54} & \textbf{5.933} & \textbf{0.045} & \textbf{331.36} & \textbf{7.109} & \textbf{0.024} & \textbf{868.63} & \textbf{4.195} & \textbf{0.023} & \textbf{352.76} & \textbf{5.392} & \textbf{0.019} & \textbf{348.51} \\
\bottomrule
\end{tabular}}
\end{table}

\begin{table}[!h]
\centering
\caption{Ablation study of Data Composition across different tasks. \textbf{UE5}: UE5 synthetic data, \textbf{Real}: real-world data.}
\label{tab:ablation_data}
\resizebox{0.75\textwidth}{!}{
\begin{tabular}{l | c c c | c c c | c c c | c c c }
\toprule
\multirow{3}{*}{Method} & \multicolumn{6}{c|}{\textbf{Trajectory-controlled}} & \multicolumn{6}{c}{\textbf{Reference-video-controlled}} \\
\cmidrule(lr){2-7} \cmidrule(lr){8-13}
 & \multicolumn{3}{c|}{T2V} & \multicolumn{3}{c|}{I2V} & \multicolumn{3}{c|}{T2V} & \multicolumn{3}{c}{I2V} \\
\cmidrule(lr){2-4} \cmidrule(lr){5-7} \cmidrule(lr){8-10} \cmidrule(lr){11-13}
 & TransErr$\downarrow$ & RotErr$\downarrow$ & FVD$\downarrow$ & TransErr$\downarrow$ & RotErr$\downarrow$ & FVD$\downarrow$ & TransErr$\downarrow$ & RotErr$\downarrow$ & FVD$\downarrow$ & TransErr$\downarrow$ & RotErr$\downarrow$ & FVD$\downarrow$ \\
\midrule
w/o UE5 & 3.105 & 0.050 & 915.22 & 4.512 & 0.058 & 350.14 & 8.845 & 0.035 & 890.56 & 5.321 & 0.030 & 365.21 \\
w/o Real & 2.152 & 0.036 & 1150.45 & 3.481 & 0.045 & 485.62 & 7.315 & 0.026 & 1085.24 & 4.352 & 0.025 & 492.35 \\
\midrule
\textbf{Ours} & \textbf{2.064} & \textbf{0.034} & \textbf{893.95} & \textbf{3.355} & \textbf{0.043} & \textbf{333.54} & \textbf{7.109} & \textbf{0.024} & \textbf{868.63} & \textbf{4.195} & \textbf{0.023} & \textbf{352.76} \\
\bottomrule
\end{tabular}}
\vspace{-1.5em}
\end{table}

\begin{figure}[!h]
  \centering
  \includegraphics[width=1\linewidth]{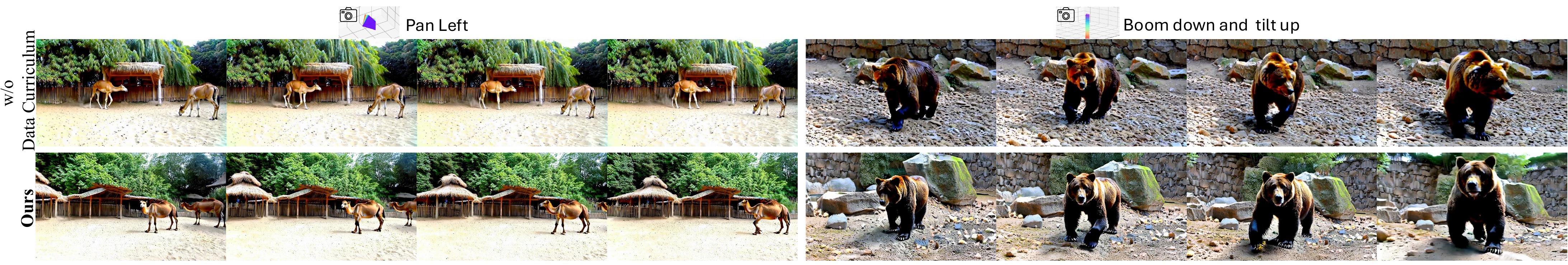}
  \vspace{-1.5em}
  
  \centerline{\tiny \textbf{(a)} w/o DC on Trajectory-controlled T2V}
  
  \includegraphics[width=1\linewidth]{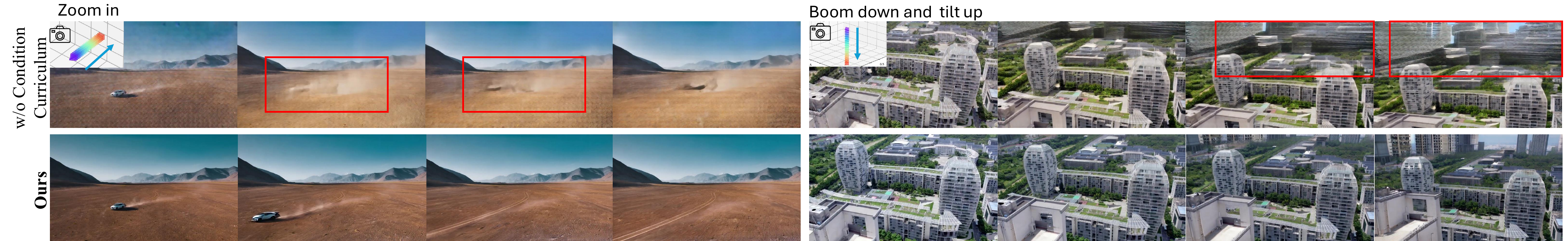}
  \vspace{-1.5em}
  
  \centerline{\tiny \textbf{(b)} w/o CC on Trajectory-controlled I2V}
  
  \includegraphics[width=1\linewidth]{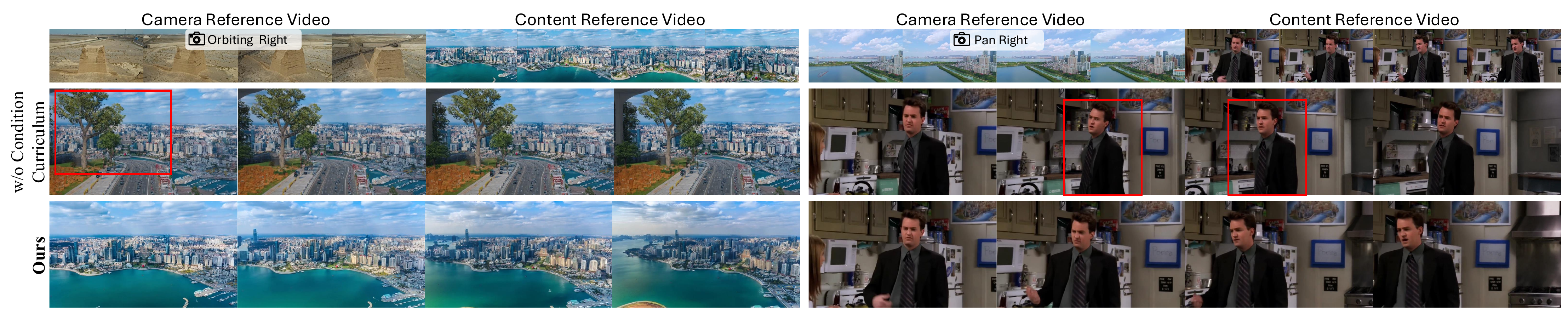}
  \vspace{-1.5em}
  
  \centerline{\tiny \textbf{(c)} w/o CC on Reference-video-controlled V2V}
  
  \includegraphics[width=1\linewidth]{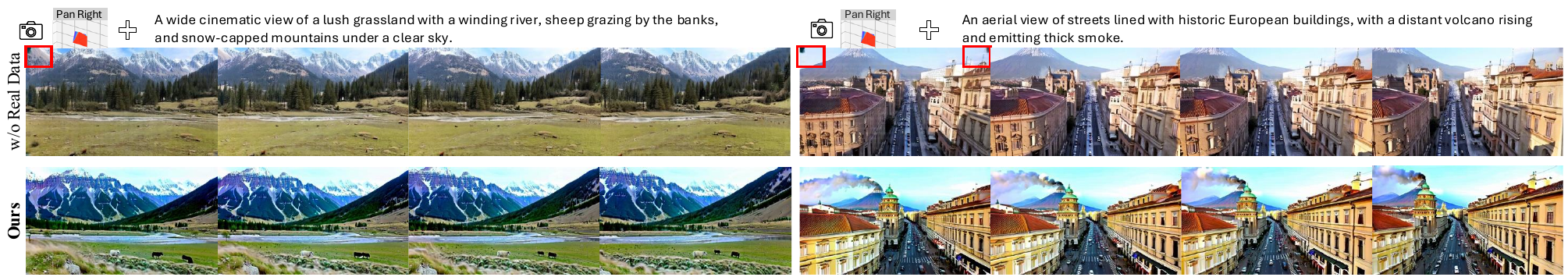}
  \vspace{-1.5em}
  
  \centerline{\tiny \textbf{(d)} w/o Real data on Trajectory-controlled T2V}
  
  \vspace{-0.5em}
  \caption{\textbf{Qualitative ablation.} \textbf{(a)} Removing DC leads to failed camera motions and degraded visual quality. \textbf{(b)(c)} Omitting CC results in structural distortions in I2V and content hallucinations in V2V (red boxes). \textbf{(d)} Training solely on synthetic data ensures precise control but lacks photorealism.}
  \label{fig:ablation_visual}
  \vspace{-2em}
\end{figure}

\vspace{-0.1em}
\subsection{Text Condition}
\vspace{-0.1em}
As shown in Fig.~\ref{fig:experiment1}, text-controlled generation provides semantic-level camera guidance by prepending motion instructions to the input caption. As shown in Tab.~\ref{tab:Text-controlled}, foundational video models (e.g., AnimateDiff~\cite{guo2023animatediff} and Wan2.2~\cite{wan2025wan}) struggle to execute specific camera commands. Even the specialized Wan2.2-Fun-Camera model can only perform simple camera movements (e.g., pan and tilt) with suboptimal precision. As further visualized in Fig.~\ref{fig:experiment4}, Wan2.2-Fun-Camera also suffers from severe content hallucination and structural artifacts (e.g., semantic confusion in animals and static volcanic smoke). In comparison, OmniCamera supports complex 3D spatial movements and outperforms all baselines in motion accuracy by a large margin while maintaining competitive visual fidelity. Additionally, OmniCamera uniquely enables \emph{text-guided} camera motion editing on existing videos (V2V), a capability under-explored in prior models.

\vspace{-0.1em}
\subsection{Reference-Video Condition}
\vspace{-0.1em}
As shown in Fig.~\ref{fig:experiment3} and Tab.~\ref{tab:Video-controlled}, OmniCamera achieves accurate reference-video-controlled camera motion across T2V, I2V, and V2V tasks. Compared with CamCloneMaster~\cite{luo2025camclonemaster} (Fig.~\ref{fig:experiment4}), OmniCamera yields significantly more accurate camera motion, superior content preservation, and fewer visual artifacts. This strong performance is directly attributed to the proposed dual-level curriculum training strategy and our superior data composition, which combines precise UE5 synthetic data with diverse real-world video pairs.


\vspace{-0.1em}
\subsection{Trajectory Condition}
\vspace{-0.1em}
As shown in Fig.~\ref{fig:experiment2} and Tab.~\ref{tab:Trajectory-controlled}, OmniCamera achieves highly accurate 3D trajectory control across T2V, I2V, and V2V tasks. Compared with state-of-the-art methods such as ReCamMaster~\cite{bai2025recammaster} (Fig.~\ref{fig:experiment4}), our approach exhibits superior motion precision with fewer visual artifacts. Benefiting from our {OmniCAM} dataset and dual-level curriculum co-training, OmniCamera closely follows diverse trajectories and generalizes to complex paths (e.g., triangular and polyline paths).


\vspace{-0.1em}
\subsection{Combining Control Modalities and Conflict Analysis}
\vspace{-0.1em}
Multiple control modalities can be combined to achieve composite camera effects. As shown in Fig.~\ref{fig:compose}, distinct conditions can be seamlessly integrated; for instance, a semantic text command (e.g., ``Pan right'') can be combined with a 3D trajectory (e.g., dolly out), producing a fused motion in the generated video. 
We further test potential modality conflicts by providing contradictory motions in pairwise combinations of trajectory, reference video, and text conditions. As shown in Fig.~\ref{fig:mode_conflict}, we observe a consistent dominance hierarchy: \textbf{Trajectory $>$ Reference Video $>$ Text}, indicating that conditions with stronger geometric explicitness naturally override weaker, semantic-level ones. For instance, when the input condition combines a ``pan left'' trajectory with a ``pan right'' reference video, the output strictly follows the ``pan left'' camera motion dictated by the trajectory. Similarly, when combining a ``pan right'' reference video with a ``pan left'' text prompt, the output adheres exclusively to the ``pan right'' motion dictated by the video.



\vspace{-0.1em}
\subsection{Ablation Study}
\vspace{-0.1em}
We perform ablation studies to analyze our dual-level curriculum, including the \emph{condition-level curriculum} (\textbf{CC}) and the \emph{data-level curriculum} (\textbf{DC}), as well as the contribution of OmniCAM's hybrid data sources by ablating either UE5 synthetic videos or curated real-world videos.

\noindent \textbf{Effect of Dual-level Curriculum.}
Tab.~\ref{tab:ablation_curriculum} validates the necessity of our dual-level curriculum. \textit{w/o CC} jointly trains all tasks from scratch; this naive mixing causes representation conflicts that fail to disentangle camera and content conditions. According to the qualitative results in Fig.~\ref{fig:ablation_visual}(b)(c), removing CC leads to severe visual degradation, including frame deterioration (e.g., in the car scene) and content hallucination (e.g., erroneous trees). Furthermore, according to the qualitative results \textit{w/o DC} in Fig.~\ref{fig:ablation_visual}(a), it hinders the learning of precise pose control, e.g., missing the ``pan left'' in the camel scene and ``boom down, tilt up'' in the bear scene, whereas ours strictly follows the desired trajectories.

\noindent \textbf{Effect of Data Composition.}
Tab.~\ref{tab:ablation_data} explores the effect of data sources by removing UE5 data (\textit{w/o UE5}) or real data (\textit{w/o Real}). We observe that UE5 data mainly contributes to camera controllability, whereas real data is crucial for closing the domain gap and recovering photorealism. According to the qualitative results of removing real data in Fig.~\ref{fig:ablation_visual}(d), relying solely on UE5 data (\textit{w/o Real}) results in a noticeable drop in visual realism and introduces erroneous artifacts such as grids in red boxes. In contrast, training with both data sources seamlessly preserves the learned camera control while yielding high-fidelity appearances.

\noindent \textbf{Further ablations.}
Please refer to the supplementary materials for additional ablations on 3D Condition RoPE and dual-condition CFG.

\section{Conclusion}
This paper presents \textbf{OmniCamera}, a unified multi-task video generation framework that supports text, trajectory, and reference-video camera control across T2V, I2V, and V2V. We construct the \textbf{OmniCAM} dataset that combines high-precision synthetic trajectories with real-world videos, and propose a dual-level curriculum co-training strategy for stable multi-condition learning. We further introduce 3D Condition RoPE and a dual-condition CFG to mitigate condition conflicts and improve camera controllability. Extensive experiments demonstrate improved controllability with competitive visual quality across tasks.

\noindent \textbf{Limitation.} OmniCamera successfully explores and demonstrates the feasibility of decoupling the observation perspective (camera) and scene content in video generation. However, while we unify the most common representations for these two dimensions, our current framework does not yet accommodate finer-grained controls, such as multiple reference images or localized motion guidance. We leave these extensions for future work.


\bibliographystyle{splncs04}
\bibliography{main}

@String(CVPR= {IEEE Conf. Comput. Vis. Pattern Recog.})

@String(CVPR  = {CVPR})

@article{he2024cameractrl,
  title={Cameractrl: Enabling camera control for text-to-video generation},
  author={He, Hao and Xu, Yinghao and Guo, Yuwei and Wetzstein, Gordon and Dai, Bo and Li, Hongsheng and Yang, Ceyuan},
  journal={arXiv preprint arXiv:2404.02101},
  year={2024}
}

@article{gao2025seedance,
  title={Seedance 1.0: Exploring the Boundaries of Video Generation Models},
  author={Gao, Yu and Guo, Haoyuan and Hoang, Tuyen and Huang, Weilin and Jiang, Lu and Kong, Fangyuan and Li, Huixia and Li, Jiashi and Li, Liang and Li, Xiaojie and others},
  journal={arXiv preprint arXiv:2506.09113},
  year={2025}
}

@article{wan2025wan,
  title={Wan: Open and advanced large-scale video generative models},
  author={Wan, Team and Wang, Ang and Ai, Baole and Wen, Bin and Mao, Chaojie and Xie, Chen-Wei and Chen, Di and Yu, Feiwu and Zhao, Haiming and Yang, Jianxiao and others},
  journal={arXiv preprint arXiv:2503.20314},
  year={2025}
}

@article{hou2024training,
  title={Training-free camera control for video generation},
  author={Hou, Chen and Chen, Zhibo},
  journal={arXiv preprint arXiv:2406.10126},
  year={2024}
}

@article{li2025realcam,
  title={Realcam-i2v: Real-world image-to-video generation with interactive complex camera control},
  author={Li, Teng and Zheng, Guangcong and Jiang, Rui and Zhan, Shuigen and Wu, Tao and Lu, Yehao and Lin, Yining and Deng, Chuanyun and Xiong, Yepan and Chen, Min and others},
  journal={arXiv preprint arXiv:2502.10059},
  year={2025}
}

@article{luo2025camclonemaster,
  title={CamCloneMaster: Enabling Reference-based Camera Control for Video Generation},
  author={Luo, Yawen and Bai, Jianhong and Shi, Xiaoyu and Xia, Menghan and Wang, Xintao and Wan, Pengfei and Zhang, Di and Gai, Kun and Xue, Tianfan},
  journal={arXiv preprint arXiv:2506.03140},
  year={2025}
}

@inproceedings{gu2025diffusion,
  title={Diffusion as shader: 3d-aware video diffusion for versatile video generation control},
  author={Gu, Zekai and Yan, Rui and Lu, Jiahao and Li, Peng and Dou, Zhiyang and Si, Chenyang and Dong, Zhen and Liu, Qifeng and Lin, Cheng and Liu, Ziwei and others},
  booktitle={Proceedings of the Special Interest Group on Computer Graphics and Interactive Techniques Conference Conference Papers},
  year={2025}
}

@inproceedings{jin2025flovd,
  title={Flovd: Optical flow meets video diffusion model for enhanced camera-controlled video synthesis},
  author={Jin, Wonjoon and Dai, Qi and Luo, Chong and Baek, Seung-Hwan and Cho, Sunghyun},
  booktitle={Proceedings of the Computer Vision and Pattern Recognition Conference},
  year={2025}
}

@article{bai2025recammaster,
  title={Recammaster: Camera-controlled generative rendering from a single video},
  author={Bai, Jianhong and Xia, Menghan and Fu, Xiao and Wang, Xintao and Mu, Lianrui and Cao, Jinwen and Liu, Zuozhu and Hu, Haoji and Bai, Xiang and Wan, Pengfei and others},
  year={2025}
}

@inproceedings{bahmani2025ac3d,
  title={Ac3d: Analyzing and improving 3d camera control in video diffusion transformers},
  author={Bahmani, Sherwin and Skorokhodov, Ivan and Qian, Guocheng and Siarohin, Aliaksandr and Menapace, Willi and Tagliasacchi, Andrea and Lindell, David B and Tulyakov, Sergey},
  booktitle={Proceedings of the Computer Vision and Pattern Recognition Conference},
  year={2025}
}

@article{lipman2022flow,
  title={Flow matching for generative modeling},
  author={Lipman, Yaron and Chen, Ricky TQ and Ben-Hamu, Heli and Nickel, Maximilian and Le, Matt},
  journal={arXiv preprint arXiv:2210.02747},
  year={2022}
}

@article{he2025cameractrl,
  title={{Cameractrl ii}: Dynamic scene exploration via camera-controlled video diffusion models},
  author={He, Hao and Yang, Ceyuan and Lin, Shanchuan and Xu, Yinghao and Wei, Meng and Gui, Liangke and Zhao, Qi and Wetzstein, Gordon and Jiang, Lu and Li, Hongsheng},
  journal={arXiv preprint arXiv:2503.10592},
  year={2025}
}

@article{guo2023animatediff,
  title={Animatediff: Animate your personalized text-to-image diffusion models without specific tuning},
  author={Guo, Yuwei and Yang, Ceyuan and Rao, Anyi and Liang, Zhengyang and Wang, Yaohui and Qiao, Yu and Agrawala, Maneesh and Lin, Dahua and Dai, Bo},
  journal={arXiv preprint arXiv:2307.04725},
  year={2023}
}

@inproceedings{li2025megasam,
  title={{MegaSaM}: Accurate, fast and robust structure and motion from casual dynamic videos},
  author={Li, Zhengqi and Tucker, Richard and Cole, Forrester and Wang, Qianqian and Jin, Linyi and Ye, Vickie and Kanazawa, Angjoo and Holynski, Aleksander and Snavely, Noah},
  booktitle={CVPR},
  year={2025}
}

@article{feng2024i2vcontrol,
  title={I2vcontrol-camera: Precise video camera control with adjustable motion strength},
  author={Feng, Wanquan and Liu, Jiawei and Tu, Pengqi and Qi, Tianhao and Sun, Mingzhen and Ma, Tianxiang and Zhao, Songtao and Zhou, Siyu and He, Qian},
  journal={arXiv preprint arXiv:2411.06525},
  year={2024}
}

@article{bai2024syncammaster,
  title={Syncammaster: Synchronizing multi-camera video generation from diverse viewpoints},
  author={Bai, Jianhong and Xia, Menghan and Wang, Xintao and Yuan, Ziyang and Fu, Xiao and Liu, Zuozhu and Hu, Haoji and Wan, Pengfei and Zhang, Di},
  journal={arXiv preprint arXiv:2412.07760},
  year={2024}
}

@inproceedings{ren2025gen3c,
  title={Gen3c: 3d-informed world-consistent video generation with precise camera control},
  author={Ren, Xuanchi and Shen, Tianchang and Huang, Jiahui and Ling, Huan and Lu, Yifan and Nimier-David, Merlin and M{\"u}ller, Thomas and Keller, Alexander and Fidler, Sanja and Gao, Jun},
  booktitle={Proceedings of the Computer Vision and Pattern Recognition Conference},
  pages={6121--6132},
  year={2025}
}

@inproceedings{zhang2025recapture,
  title={Recapture: Generative video camera controls for user-provided videos using masked video fine-tuning},
  author={Zhang, David Junhao and Paiss, Roni and Zada, Shiran and Karnad, Nikhil and Jacobs, David E and Pritch, Yael and Mosseri, Inbar and Shou, Mike Zheng and Wadhwa, Neal and Ruiz, Nataniel},
  booktitle={Proceedings of the Computer Vision and Pattern Recognition Conference},
  pages={2050--2062},
  year={2025}
}

@article{yu2024viewcrafter,
  title={Viewcrafter: Taming video diffusion models for high-fidelity novel view synthesis},
  author={Yu, Wangbo and Xing, Jinbo and Yuan, Li and Hu, Wenbo and Li, Xiaoyu and Huang, Zhipeng and Gao, Xiangjun and Wong, Tien-Tsin and Shan, Ying and Tian, Yonghong},
  journal={arXiv preprint arXiv:2409.02048},
  year={2024}
}

@article{xu2024camco,
  title={Camco: Camera-controllable 3d-consistent image-to-video generation},
  author={Xu, Dejia and Nie, Weili and Liu, Chao and Liu, Sifei and Kautz, Jan and Wang, Zhangyang and Vahdat, Arash},
  journal={arXiv preprint arXiv:2406.02509},
  year={2024}
}

@article{brooks2024video,
  title={Video generation models as world simulators},
  author={Brooks, Tim and Peebles, Bill and Holmes, Connor and DePue, Will and Guo, Yufei and Jing, Li and Schnurr, David and Taylor, Joe and Luhman, Troy and Luhman, Eric and others},
  journal={OpenAI Blog},
  volume={1},
  number={8},
  pages={1},
  year={2024}
}

@article{hong2022cogvideo,
  title={Cogvideo: Large-scale pretraining for text-to-video generation via transformers},
  author={Hong, Wenyi and Ding, Ming and Zheng, Wendi and Liu, Xinghan and Tang, Jie},
  journal={arXiv preprint arXiv:2205.15868},
  year={2022}
}

@article{ma2025latte,
  title={Latte: Latent diffusion transformer for video generation},
  author={Ma, Xin and Wang, Yaohui and Chen, Xinyuan and Jia, Gengyun and Liu, Ziwei and Li, Yuan-Fang and Chen, Cunjian and Qiao, Yu},
  journal={Transactions on Machine Learning Research},
  year={2025}
}

@article{blattmann2023stable,
  title={Stable video diffusion: Scaling latent video diffusion models to large datasets},
  author={Blattmann, Andreas and Dockhorn, Tim and Kulal, Sumith and Mendelevitch, Daniel and Kilian, Maciej and Lorenz, Dominik and Levi, Yam and English, Zion and Voleti, Vikram and Letts, Adam and others},
  journal={arXiv preprint arXiv:2311.15127},
  year={2023}
}

@inproceedings{xing2024dynamicrafter,
  title={Dynamicrafter: Animating open-domain images with video diffusion priors},
  author={Xing, Jinbo and Xia, Menghan and Zhang, Yong and Chen, Haoxin and Yu, Wangbo and Liu, Hanyuan and Liu, Gongye and Wang, Xintao and Shan, Ying and Wong, Tien-Tsin},
  booktitle={European Conference on Computer Vision},
  pages={399--417},
  year={2024},
  organization={Springer}
}

@article{yang2024cogvideox,
  title={CogVideoX: Text-to-Video Diffusion Models with An Expert Transformer},
  author={Yang, Zhuoyi and Teng, Jiayan and Zheng, Wendi and Ding, Ming and Huang, Shiyu and Xu, Jiazheng and Yang, Yuanming and Hong, Wenyi and Zhang, Xiaohan and Feng, Guanyu and others},
  journal={arXiv preprint arXiv:2408.06072},
  year={2024}
}

@inproceedings{esser2024scaling,
  title={Scaling rectified flow transformers for high-resolution image synthesis},
  author={Esser, Patrick and Kulal, Sumith and Blattmann, Andreas and Entezari, Rahim and M{\"u}ller, Jonas and Saini, Harry and Levi, Yam and Lorenz, Dominik and Sauer, Axel and Boesel, Frederic and others},
  booktitle={Forty-first international conference on machine learning},
  year={2024}
}

@misc{flux2024,
    author={Black Forest Labs},
    title={FLUX},
    year={2024},
    howpublished={\url{https://github.com/black-forest-labs/flux}},
}

@inproceedings{peebles2023scalable,
  title={Scalable diffusion models with transformers},
  author={Peebles, William and Xie, Saining},
  booktitle={Proceedings of the IEEE/CVF international conference on computer vision},
  pages={4195--4205},
  year={2023}
}

@inproceedings{yang2024depth,
  title={Depth anything: Unleashing the power of large-scale unlabeled data},
  author={Yang, Lihe and Kang, Bingyi and Huang, Zilong and Xu, Xiaogang and Feng, Jiashi and Zhao, Hengshuang},
  booktitle={Proceedings of the IEEE/CVF conference on computer vision and pattern recognition},
  pages={10371--10381},
  year={2024}
}

@article{bian2025gs,
  title={Gs-dit: Advancing video generation with pseudo 4d gaussian fields through efficient dense 3d point tracking},
  author={Bian, Weikang and Huang, Zhaoyang and Shi, Xiaoyu and Li, Yijin and Wang, Fu-Yun and Li, Hongsheng},
  journal={arXiv preprint arXiv:2501.02690},
  year={2025}
}

@inproceedings{van2024generative,
  title={Generative camera dolly: Extreme monocular dynamic novel view synthesis},
  author={Van Hoorick, Basile and Wu, Rundi and Ozguroglu, Ege and Sargent, Kyle and Liu, Ruoshi and Tokmakov, Pavel and Dave, Achal and Zheng, Changxi and Vondrick, Carl},
  booktitle={European Conference on Computer Vision},
  pages={313--331},
  year={2024},
  organization={Springer}
}

@article{yu2025trajectorycrafter,
  title={Trajectorycrafter: Redirecting camera trajectory for monocular videos via diffusion models},
  author={YU, Mark and Hu, Wenbo and Xing, Jinbo and Shan, Ying},
  journal={arXiv preprint arXiv:2503.05638},
  year={2025}
}

@article{zhou2018stereo,
  title={Stereo magnification: Learning view synthesis using multiplane images},
  author={Zhou, Tinghui and Tucker, Richard and Flynn, John and Fyffe, Graham and Snavely, Noah},
  journal={arXiv preprint arXiv:1805.09817},
  year={2018}
}

@inproceedings{dai2017scannet,
    title={ScanNet: Richly-annotated 3D Reconstructions of Indoor Scenes},
    author={Dai, Angela and Chang, Angel X. and Savva, Manolis and Halber, Maciej and Funkhouser, Thomas and Nie{\ss}ner, Matthias},
    booktitle = CVPR,
    year = {2017}
}

@article{Matterport3D,
  title={Matterport3D: Learning from RGB-D Data in Indoor Environments},
  author={Chang, Angel and Dai, Angela and Funkhouser, Thomas and Halber, Maciej and Niessner, Matthias and Savva, Manolis and Song, Shuran and Zeng, Andy and Zhang, Yinda},
  journal={International Conference on 3D Vision (3DV)},
  year={2017}
}

@inproceedings{ling2024dl3dv,
  title={Dl3dv-10k: A large-scale scene dataset for deep learning-based 3d vision},
  author={Ling, Lu and Sheng, Yichen and Tu, Zhi and Zhao, Wentian and Xin, Cheng and Wan, Kun and Yu, Lantao and Guo, Qianyu and Yu, Zixun and Lu, Yawen and others},
  booktitle=CVPR,
  year={2024}
}

@inproceedings{
dehghan2021arkitscenes,
title={{ARK}itScenes - A Diverse Real-World Dataset for 3D Indoor Scene Understanding Using Mobile {RGB}-D Data},
author={Gilad Baruch and Zhuoyuan Chen and Afshin Dehghan and Tal Dimry and Yuri Feigin and Peter Fu and Thomas Gebauer and Brandon Joffe and Daniel Kurz and Arik Schwartz and Elad Shulman},
booktitle={Thirty-fifth Conference on Neural Information Processing Systems Datasets and Benchmarks Track (Round 1)},
year={2021},
url={https://openreview.net/forum?id=tjZjv_qh_CE}
}

@article{ye2025unic,
  title={Unic: Unified in-context video editing},
  author={Ye, Zixuan and He, Xuanhua and Liu, Quande and Wang, Qiulin and Wang, Xintao and Wan, Pengfei and Zhang, Di and Gai, Kun and Chen, Qifeng and Luo, Wenhan},
  journal={arXiv preprint arXiv:2506.04216},
  year={2025}
}

@article{cai2025omnivcus,
  title={Omnivcus: Feedforward subject-driven video customization with multimodal control conditions},
  author={Cai, Yuanhao and Zhang, He and Chen, Xi and Xing, Jinbo and Hu, Yiwei and Zhou, Yuqian and Zhang, Kai and Zhang, Zhifei and Kim, Soo Ye and Wang, Tianyu and others},
  journal={arXiv preprint arXiv:2506.23361},
  year={2025}
}
\end{document}